\documentclass[fleqn,10pt]{wlscirep}
\usepackage{float}

\title{Detecting and classifying lesions in mammograms with Deep Learning}
\author[1*]{Dezső Ribli}
\author[2]{Anna Horváth}
\author[3]{Zsuzsa Unger}
\author[4]{Péter Pollner}
\author[1]{István Csabai}

\affil[1]{Department of Physics of Complex Systems, Eötvös Loránd University, Budapest}
\affil[2]{3rd Department of Internal Medicine, Semmelweis University, Budapest}
\affil[3]{Department of Radiology, Semmelweis University, Budapest}
\affil[4]{MTA-ELTE Statistical and Biological Physics Research Group, Hungarian Academy of Sciences, Budapest}

\affil[*]{dkrib@caesar.elte.hu}

\begin{abstract}

In the last two decades Computer Aided Diagnostics (CAD) systems were developed to help radiologists analyze screening mammograms. 
The benefits of current CAD technologies appear to be contradictory and they should be improved to be ultimately considered useful.
Since 2012 deep convolutional neural networks (CNN) have been a tremendous success in image recognition, reaching human performance.
These methods have  greatly surpassed the traditional approaches,  which are similar to currently used CAD solutions.
Deep CNN-s have the potential to revolutionize medical image analysis.
We propose a CAD system based on one of the most successful object detection frameworks, Faster R-CNN.
The system detects and classifies malignant or benign lesions on a mammogram without any human intervention.
The proposed method sets the state of the art classification performance on the public INbreast database, AUC $= 0.95$.
The approach described here has achieved the 2nd place in the Digital Mammography DREAM Challenge with  AUC $= 0.85 $.
When used as a detector, the system reaches high sensitivity with very few false positive marks per image on the INbreast dataset.
Source code, the trained model and an OsiriX plugin are published online at \url{https://github.com/riblidezso/frcnn_cad} .

\end{abstract}
\begin{document}

\flushbottom
\maketitle

\thispagestyle{empty}

\section*{Introduction}

%screening
\subsection*{Screening mammography}

%breast cancer and screening in general
Breast cancer is the most common cancer in women and it is the main cause of death from cancer among women in the world \cite{ferlay2010global}.
Screening mammography was shown to reduce breast cancer mortality by 38–48\% among participants \cite{broeders2012impact}.
In the EU 25 of the 28 member states are planning, piloting or implementing screening programs to diagnose and treat breast cancer in an early stage \cite{ponti2017cancer}.
During a standard mammographic screening examination, X-ray images are captured from 2 angles of each breast.
These images are inspected for malignant lesions by one or two experienced radiologists.
Suspicious cases are called back for further diagnostic evaluation.

%humans
Screening mammograms are evaluated by human readers.
The reading process is monotonous, tiring, lengthy, costly and most importantly prone to errors.
%missed cancers could be found
Multiple studies have shown that 20-30\% of the diagnosed cancers could be found retrospectively on the previous negative screening exam by blinded reviewers
\cite{bae2014breast,
bird1992analysis,
birdwell2001mammographic,
harvey1993previous,
hoff2012breast,
martin1979breast}.
The problem of missed cancers still persists despite modern full field digital mammography (FFDM)
\cite{bae2014breast,hoff2012breast}.
% sensitivity specificity inter radiologist variance
The sensitivity  and specificity of screening  mammography is reported to be between 77-87\% and 89-97\% respectively.
These metrics describe the average performance of readers, and there is substantial variance in the performance of individual physicians, with reported false positive rates between 1-29\%, and sensitivities between 29-97\%
\cite{
banks2004influence,
lehman2015diagnostic,
smith2005physician}.
%double and multiple reading, room for improvement
Double reading was found to improve the performance of mammographic evaluation and it had been implemented in many countries \cite{blanks1998comparison}.
Multiple reading can further improve diagnostic performance up to more than 10 readers, proving that there is room for improvement in mammogram evaluation beyond double reading
\cite{karssemeijer2004effect}.

%cad
\subsection*{Computer-aided detection in mammographic screening}

%cad is widespread
Computer-aided detection (CAD) solutions were developed to help radiologists in reading mammograms.
These programs usually analyze a mammogram and mark the suspicious regions, which should be reviewed by the radiologist \cite{christoyianni2002computer}.
The technology was approved by FDA and had spread quickly. 
By 2008, in the US, 74\% of all screening mammograms in the Medicare population were interpreted with CAD, the cost of CAD usage is over \$400 million a year.
\cite{lehman2015diagnostic}.

%controversial benefits
The benefits of using CAD are controversial.
Initially several studies have shown promising results with CAD
\cite{birdwell2001mammographic, brem2003improvement, 
ciatto2003comparison, freer2001screening, morton2006screening, warren2000potential}.
A large clinical trial in the United Kingdom has shown that single reading with CAD assistance has similar performance to double reading.
\cite{gilbert2008single}
However, in the last decade multiple studies concluded that currently used CAD technologies do not improve the performance of radiologists in everyday practice in the United States.
\cite{fenton2007influence,fenton2011effectiveness, lehman2015diagnostic}.
These controversial results indicate that CAD systems need to be improved before radiologists can ultimately benefit from using the technology in everyday practice.

%better cad
Currently used CAD approaches are based on describing the X-ray image with meticulously designed hand crafted features, and machine learning for classification on top of these features
\cite{christoyianni2002computer,
imagechecker10manual,
hupse2009use,
hupse2013standalone,
kooi2017large} .
In the field of computer vision, since 2012, deep convolutional neural networks (CNN) have significantly outperformed these traditional methods\cite{krizhevsky2012imagenet}.
Deep CNN-s have reached or even surpassed human performance in image classification and object detection.
\cite{he2015delving}.
These models have tremendous potential in medical image analysis.
%deep learning in cad
Several studies have attempted to apply Deep Learning to analyze mammograms \cite{becker2017deep,dhungel2017fully,kooi2017large,lotter2017multi}, but the problem is still far from being solved.

%DM challenge
\subsection*{The Digital Mammography DREAM Challenge}

The Digital Mammography DREAM Challenge (DM challenge) \cite{dmchallenge,trister2017will} asked participants to write algorithms which can predict whether a breast in a screening mammography exam will be diagnosed with cancer.
The dataset consisted of 86000 exams, with no pixel level annotation, only a binary label indicating if breast cancer was diagnosed in the next 12 months after the exam.
Each side of the breasts were treated as a separate case that we will call breast-level prediction in this paper.
The participants had to upload their programs to a secure cloud platform, and they were not able to download or view the images, neither interact with their program during training or testing.
The DM challenge provided an excellent opportunity to compare the performance of competing methods in a controlled and fair way instead of self-reported evaluations on different or proprietary datasets.

\section{Material and methods}

\subsection*{Data}

%datasets
Mammograms with pixel level annotations were needed to train a lesion detector and test the classification and localization performance.
We have trained the model on the public Digital Database for Screening Mammography  (DDSM) \cite{heath2000digital} and a dataset from the Semmelweis University in Budapest, and tested it on the public INbreast \cite{moreira2012inbreast} dataset.
%both malignant and benign in training
The images used for training contain either histologically proven cancers or benign lesions which were recalled for further examinations, but later turned out to be nonmalignant.
We expect that training with both kinds of lesions helps our model to find more lesions of interest, and differentiate between malignant and benign examples.

%ddsm
The DDSM  dataset contains 2620 digitized film-screen screening mammography exams, with pixel-level ground truth annotation of lesions. 
Cancerous lesions have histological proof.
We have only used the DDSM database for training our model and not evaluating it.
The quality of digitized film-screen mammograms is not as good as full field digital mammograms therefore, evaluation on these cases is not relevant.
We have converted the lossless jpeg images to png format, mapped the pixel values to optical density using the calibration functions from the DDSM website, and rescaled the pixel values to the 0-255 range.

%se
The dataset from the Department of Radiology at the Semmelweis University in Budapest, Hungary contains 847 FFDM images of 214 exams from 174 patients, recorded with a Hologic LORAD Selenia device.
Institutional board approval was obtained for the dataset.
This dataset was not available for the full period of the DM challenge, it is used only for improvement in the second stage of the DM challenge, after the pixel level annotation by the authors.

%inbreast
The INbreast dataset contains 115 FFDM cases with pixel-level ground truth annotations, and histological proof for cancers
\cite{moreira2012inbreast}.
We have adapted the INbreast pixel level annotations to suit our testing scenario. 
We have ignored all benign annotations, and converted the malignant lesion annotations to bounding boxes.
We have excluded 8 exams which had other findings, artifacts, previous surgeries or ambiguous pathological outcome.
The images have low contrast therefore, we have adjusted the window of the pixel levels.
The pixel values were clipped to be minimum 500 pixel lower and maximum 800 pixels higher than the mode of the pixel value distribution (excluding the background) and were rescaled to the 0-255 range.

\subsection*{Data Availability}

The DDSM dataset is available online at \url{http://marathon.csee.usf.edu/Mammography/Database.html}.\\
The INBreast dataset can be requested online at \url{http://medicalresearch.inescporto.pt/breastresearch/index.php/Get_INbreast_Database}.\\
For the dataset from the Semmelweis university (http://semmelweis.hu/radiologia/) restrictions apply to the availability of these data, which were used under special licence from Semmelweis University, and so are not publicly available. Data are however available from the authors upon reasonable request and permission of the Semmelweis University.

\subsection*{Methods}

\begin{figure}[H]
  \centering    \includegraphics[width=1.0\textwidth]{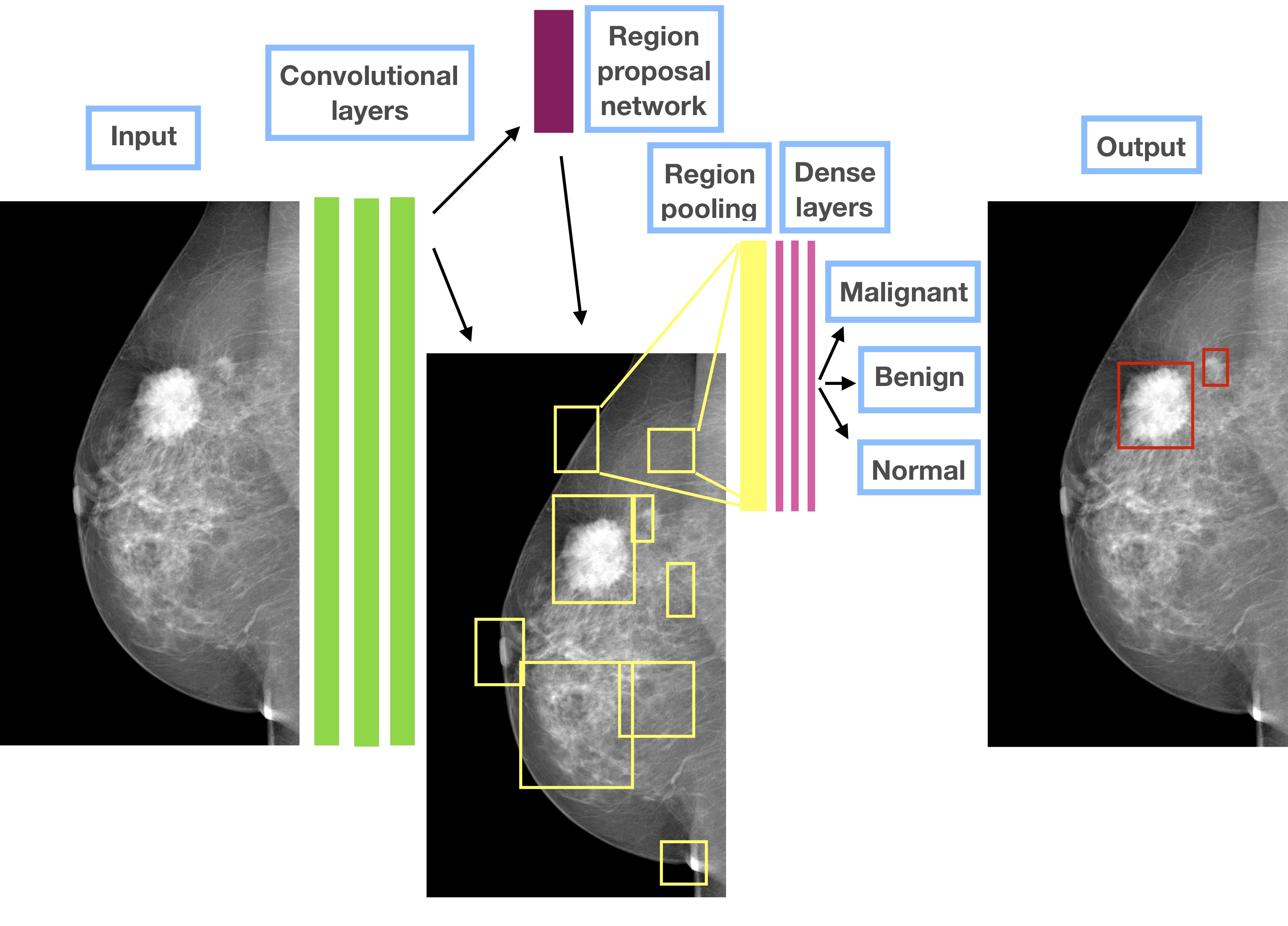}
    \caption{ The outline of the Faster R-CNN model for CAD in mammography. }
    \label{fig:frcnn_cad}
\end{figure}

%Faster rcnn and vgg16
The heart of our model is a state of the art object detection framework, Faster R-CNN  \cite{ren2015faster}.
Faster R-CNN is based on a convolutional neural network with additional components for detecting, localizing and classifying objects in an image.
% rpn
Faster R-CNN has a branch of convolutional layers, called Region Proposal Network (RPN), after the last convolutional layer of the original network, which is trained to detect and localize objects on the image, regardless of the class of the object. 
There are default detection boxes with different sizes and aspect ratios in order to find objects with varying sizes and shapes.
The highest scoring default boxes are called the region proposals for the other branch of the network.
% rcnn
The other branch of the neural network evaluates the signal coming from each proposed region of the last convolutional layer, resampled to a fix size.
% detect + bbox
Both branches try to solve a classification task to detect the presence of objects and a bounding-box regression task in order to refine the boundaries of the object present in the region.
%nms
From the detected overlapping objects, the best predictions are selected using non-maximum suppression.
Further details about Faster R-CNN can be found in the original article \cite{ren2015faster}.
An outline of the model can be seen in Fig. ~\ref{fig:frcnn_cad}.

%roc

The base CNN used in our model was a VGG16 network, which is 16 layer deep CNN \cite{simonyan2014very}.
%final output in one image
The final layer can detect 2 kinds of objects in the images, benign lesions or malignant lesions.
The model outputs a bounding box for each detected lesion, and a score, which reflects the confidence in the class of the lesion.
% whole image usage
To describe an image with one score, we take the maximum of the scores of all malignant lesions detected in the image.
For multiple images of the same breast, we take the average of the scores of individual images.
%multi model usage
For the DM challenge we have trained 2 models using shuffled training datasets.
When ensembling these models, the score of an image was the average score of the individual models.
This approach has been motivated by a previous study on independent human readers, and it has proven reasonably effective, while simple and flexible \cite{karssemeijer2004effect}.

%technical details
We have used the framework developed by the authors of Faster R-CNN \cite{ren2015faster}, which was built in the Caffe framework for deep learning \cite{jia2014caffe}.
During training we optimized the object detection part and the classifier part of the model in the same time, this is called the joint optimization \cite{ren2015faster}.
We used backpropagation and stochastic gradient descent with weight decay.
The initial model used for training has been pretrained on 1.2 million images from the ImageNet dataset \cite{simonyan2014very}.

%resolution
We have found that higher resolution yields better results, therefore the mammograms were rescaled isotropically to longer side smaller than 2100 pixels or shorter side smaller than 1700 pixels.
This resolution is close to the maximum size which fits in the memory of the graphics card used.
The aspect ratio was selected to fit the regular aspect ratio of Hologic images.
% augmentation
We applied vertical and horizontal flipping to augment the training dataset.
% rpn positive overlap setup
Mammograms contain fewer object than ordinary images, therefore negative proposals dominate minibatches.
The Intersection over Union (IoU) threshold for foreground objects in the region proposal network was relaxed from 0.7 to 0.5 to allow more positive samples in each minibatch.
Relaxation of positive examples is also supported by the fact that lesions on a mammogram have much less well defined boundaries than a car or a dog on a tradiotional image.
%final iou
The IoU threshold of the final non maximum suppression (nms) was set to $0.1$, because mammograms represent a smaller and compressed 3D space compared to ordinary images, therefore overlapping detections are expected to happen less often than in usual object detection. 
The model was trained for 40k iterations, this number was previously found to be close to optimal by testing multiple models on the DM challenge training data.
The model was trained and evaluated on an Nvidia GTX 1080Ti graphics card.
%differences in the challenge
Our final entry in the DM challenge was an ensemble of 2 models.

\section*{Results}

% roc auc for cancer classification
\subsection*{Cancer classification}

\begin{figure}[H]
  \centering    \includegraphics[width=0.8\textwidth]{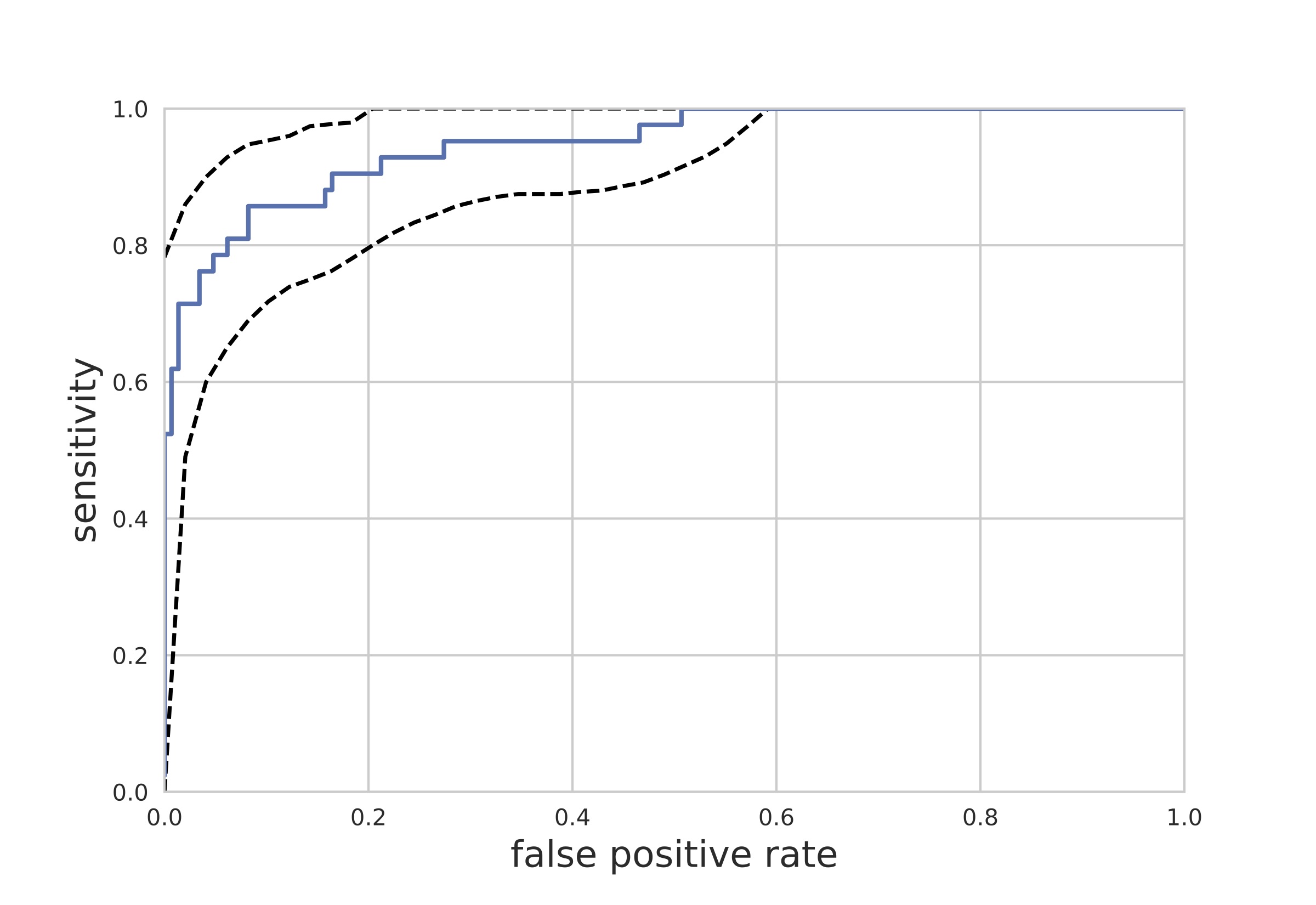}
    \caption{Classification performance. The solid blue line shows the ROC curve on the INbreast dataset on breast level, AUC $= 0.95 $, the dashed lines show the 95 percentile interval of the curve based on 10000 bootstrap samples.  }
    \label{fig:roc}
\end{figure}

%inbreast
We also evaluated the model's performance on the public INbreast dataset with the receiver operating characteristics (ROC) metric, Fig.~\ref{fig:roc}.
The INbreast dataset has many exams with only one laterality, therefore we have evaluated predictions for each breast.
The system achieved AUC $=0.95$, (95 percentile interval: 0.91 to 0.98, estimated from 10000 bootstrap samples). 
To our best knowledge this is the highest AUC score reported on the INbreast dataset with a fully automated system based on a single model.

%roc

%localized measure: froc
\subsection*{FROC analysis}

\begin{figure}[H]
  \centering    \includegraphics[width=0.8\textwidth]{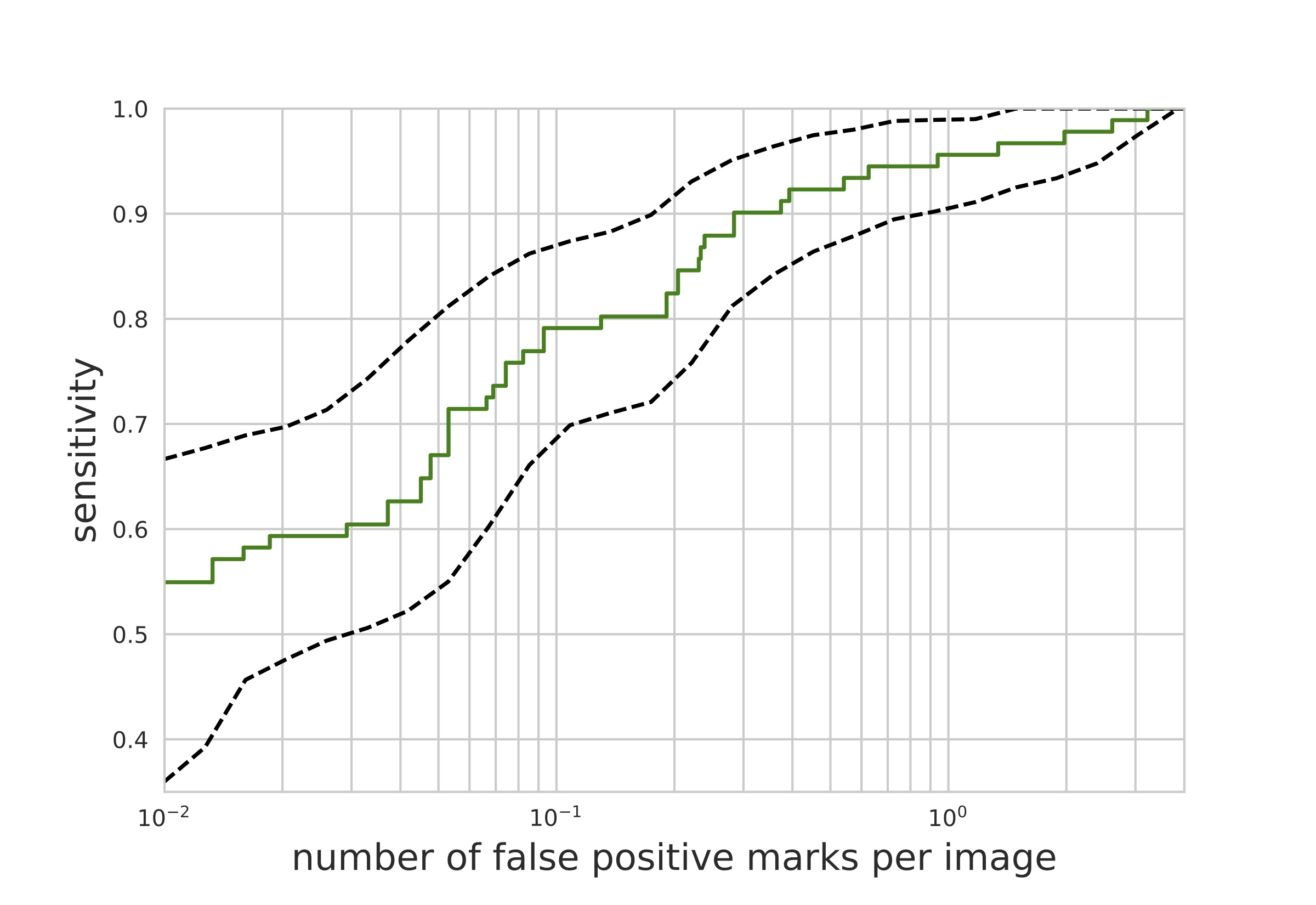}
    \caption{FROC curve on the INbreast dataset. Sensitivity is calculated on a per lesion basis. The solid curve with squares shows the results using all images, while the dashed lines show the 95 percentile interval from 10000 bootstrap samples.}
    \label{fig:froc}
\end{figure}

%froc
In order to test the model's ability to detect and accurately localize malignant lesions, we evaluated the predictions on the INbreast dataset using the Free-response ROC (FROC) curve \cite{bunch1977free}.
The FROC curve shows the sensitivity (fraction of correctly localized lesions) as a function of the number of false positive marks put on an image Fig.~\ref{fig:froc}.

%rules
A detection was considered correct if the center of the proposed lesion fell inside a ground truth box.
The same criteria is generally used when measuring the performance of currently used CAD products \cite{ellis2007evaluation,imagechecker10manual, sadaf2011performance}.
The DM challenge dataset has no lesion annotation, therefore we can not use it for an FROC analysis.

%fig

%some examples 
\subsection*{Examples}

\begin{figure}[h!]
  \centering
  \includegraphics[width=1.0\textwidth]{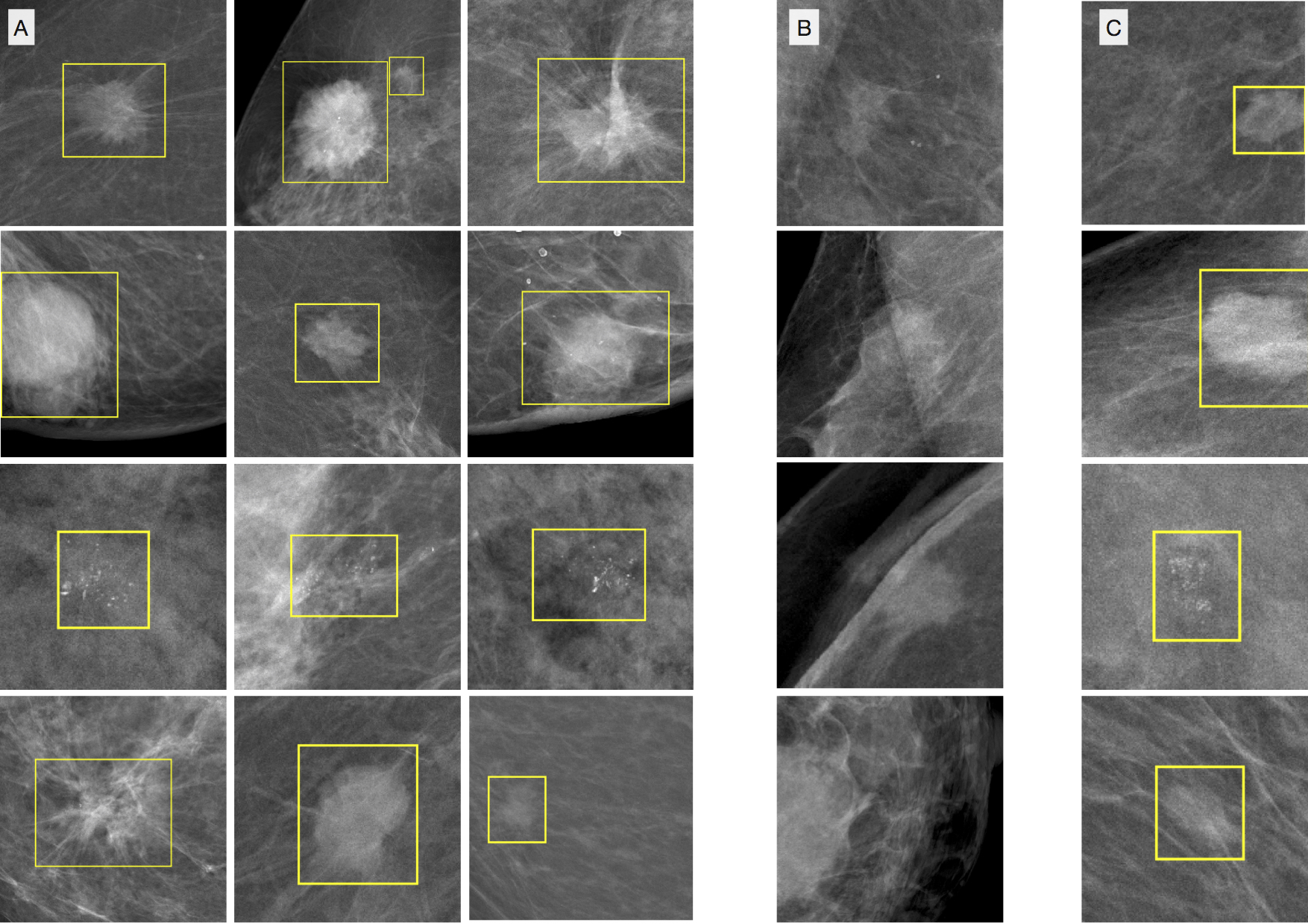}
   \caption{Detection examples: The yellow boxes show the lesion proposed by the model. The threshold for these detections was selected to be at lesion detection sensitivity = $0.9$. A: Correctly detected malignant lesions B: Missed malignant lesions C: False positive detections,  Courtesy of the Breast Research Group, INESC Porto, Portugal \cite{moreira2012inbreast}}
   \label{fig:examples}
\end{figure}

%hands on experience
To demonstrate the characteristics and errors of the detector, we have created a collection of correctly classified, false positive and missed malignant lesions of the INbreast dataset, see in Fig.~\ref{fig:examples}.
The score threshold for the examples was defined at sensitivity $=0.9$ and $0.3$ false positive marks per image.

%qualitative evaluation of example
% false positive are mostly benign lesions
After inspecting the false positive detections, we have found that most were benign masses or calcifications.
Some of these benign lesions were biopsy tested according to the case descriptions of the INbreast dataset.
% false negatives are detected with slightly higher false positive rate
While 10\% of the ground truth malignant lesions were missed at this detection threshold, these were not completely overlooked by the model.
With a score threshold which corresponds to $3$ false positive marks per image, all the lesions were correctly detected (see Fig.~\ref{fig:froc}).
Note that the exact number of false positive and true positive detections slightly varies with with different samples of images, indicated by the area in
Fig.~\ref{fig:froc}.

%fig

\section*{Discussion}

%DM 2nd place, INbreast state of the art, great detection 
We have proposed a Faster R-CNN based CAD approach, which achieved the 2nd position in the Digital Mammography DREAM Challenge with an AUC = $0.85$ score on the final validation dataset.
The competition results have proved that the method described in this article is one of the best approach for cancer classification in mammograms.
%it is a detecor inherently
Our method was the only one of the top contenders in the DM challenge which is based on the detection of malignant lesions, and whole image classification is just a trivial step from the detection task.
We think that a lesion detector is clinically much more useful than a simple classifier.
A classifier only gives a single score for a case or breast, but it is not able to locate the cancer which is essential for further diagnostic tests or treatment.

%INbreast evaluation
We have evaluated the model on the publicly available INbreast dataset.
The system is able to detect 90\% of the malignant lesions in the INbreast dataset with only 0.3 false positive marks per image.
It also sets the state of the art performance in cancer classification on the publicly available INbreast dataset.
The system uses the mammograms as the only input without any annotation or user interaction.

%CAD usage
An object detection framework developed to detect objects in ordinary images shows excellent performance.
This result indicates that lesion detection on mammograms is not very different from the regular object detection task. 
Therefore the expensive, traditional CAD solutions, which have controversial efficiency, could be replaced with the recently developed, potentially more accurate, deep learning based, open source object detection methods in the near future.
The FROC analysis results suggest that the proposed model could be applied as a perception enhancer tool, which could help radiologists to detect more cancers.

%limitation: small dataset
A limitation of our study comes from the small size of the publicly available pixel-level annotated dataset.
While the classification performance of the model has been evaluated on a large screening dataset, the detection  performance could only be evaluated on the small INbreast dataset.

\bibliography{ref}

\begin{thebibliography}{10}
\expandafter\ifx\csname url\endcsname\relax
  \def\url#1{\texttt{#1}}\fi
\expandafter\ifx\csname urlprefix\endcsname\relax\def\urlprefix{URL }\fi
\expandafter\ifx\csname doiprefix\endcsname\relax\def\doiprefix{DOI }\fi
\providecommand{\bibinfo}[2]{#2}
\providecommand{\eprint}[2][]{\url{#2}}

\bibitem{ferlay2010global}
\bibinfo{author}{Ferlay, J.}, \bibinfo{author}{H{\'e}ry, C.},
  \bibinfo{author}{Autier, P.} \& \bibinfo{author}{Sankaranarayanan, R.}
\newblock \bibinfo{title}{Global burden of breast cancer}.
\newblock In \emph{\bibinfo{booktitle}{Breast cancer epidemiology}},
  \bibinfo{pages}{1--19} (\bibinfo{publisher}{Springer}, \bibinfo{year}{2010}).

\bibitem{broeders2012impact}
\bibinfo{author}{Broeders, M.} \emph{et~al.}
\newblock \bibinfo{journal}{\bibinfo{title}{The impact of mammographic
  screening on breast cancer mortality in europe: a review of observational
  studies}}.
\newblock {\emph{\JournalTitle{Journal of medical screening}}}
  \textbf{\bibinfo{volume}{19}}, \bibinfo{pages}{14--25}
  (\bibinfo{year}{2012}).

\bibitem{ponti2017cancer}
\bibinfo{author}{Ponti, A.} \emph{et~al.}
\newblock \bibinfo{title}{Cancer screening in the european union. final report
  on the implementation of the council recommendation on cancer screening}
  (\bibinfo{year}{2017}).

\bibitem{bae2014breast}
\bibinfo{author}{Bae, M.~S.} \emph{et~al.}
\newblock \bibinfo{journal}{\bibinfo{title}{Breast cancer detected with
  screening us: reasons for nondetection at mammography}}.
\newblock {\emph{\JournalTitle{Radiology}}} \textbf{\bibinfo{volume}{270}},
  \bibinfo{pages}{369--377} (\bibinfo{year}{2014}).

\bibitem{bird1992analysis}
\bibinfo{author}{Bird, R.~E.}, \bibinfo{author}{Wallace, T.~W.} \&
  \bibinfo{author}{Yankaskas, B.~C.}
\newblock \bibinfo{journal}{\bibinfo{title}{Analysis of cancers missed at
  screening mammography.}}
\newblock {\emph{\JournalTitle{Radiology}}} \textbf{\bibinfo{volume}{184}},
  \bibinfo{pages}{613--617} (\bibinfo{year}{1992}).

\bibitem{birdwell2001mammographic}
\bibinfo{author}{Birdwell, R.~L.}, \bibinfo{author}{Ikeda, D.~M.},
  \bibinfo{author}{O’Shaughnessy, K.~F.} \& \bibinfo{author}{Sickles, E.~A.}
\newblock \bibinfo{journal}{\bibinfo{title}{Mammographic characteristics of 115
  missed cancers later detected with screening mammography and the potential
  utility of computer-aided detection 1}}.
\newblock {\emph{\JournalTitle{Radiology}}} \textbf{\bibinfo{volume}{219}},
  \bibinfo{pages}{192--202} (\bibinfo{year}{2001}).

\bibitem{harvey1993previous}
\bibinfo{author}{Harvey, J.~A.}, \bibinfo{author}{Fajardo, L.~L.} \&
  \bibinfo{author}{Innis, C.~A.}
\newblock \bibinfo{journal}{\bibinfo{title}{Previous mammograms in patients
  with impalpable breast carcinoma: retrospective vs blinded interpretation.
  1993 arrs president's award.}}
\newblock {\emph{\JournalTitle{AJR. American journal of roentgenology}}}
  \textbf{\bibinfo{volume}{161}}, \bibinfo{pages}{1167--1172}
  (\bibinfo{year}{1993}).

\bibitem{hoff2012breast}
\bibinfo{author}{Hoff, S.~R.} \emph{et~al.}
\newblock \bibinfo{journal}{\bibinfo{title}{Breast cancer: missed interval and
  screening-detected cancer at full-field digital mammography and screen-film
  mammography—results from a retrospective review}}.
\newblock {\emph{\JournalTitle{Radiology}}} \textbf{\bibinfo{volume}{264}},
  \bibinfo{pages}{378--386} (\bibinfo{year}{2012}).

\bibitem{martin1979breast}
\bibinfo{author}{Martin, J.~E.}, \bibinfo{author}{Moskowitz, M.} \&
  \bibinfo{author}{Milbrath, J.~R.}
\newblock \bibinfo{journal}{\bibinfo{title}{Breast cancer missed by
  mammography}}.
\newblock {\emph{\JournalTitle{American Journal of Roentgenology}}}
  \textbf{\bibinfo{volume}{132}}, \bibinfo{pages}{737--739}
  (\bibinfo{year}{1979}).

\bibitem{banks2004influence}
\bibinfo{author}{Banks, E.} \emph{et~al.}
\newblock \bibinfo{journal}{\bibinfo{title}{Influence of personal
  characteristics of individual women on sensitivity and specificity of
  mammography in the million women study: cohort study}}.
\newblock {\emph{\JournalTitle{Bmj}}} \textbf{\bibinfo{volume}{329}},
  \bibinfo{pages}{477} (\bibinfo{year}{2004}).

\bibitem{lehman2015diagnostic}
\bibinfo{author}{Lehman, C.~D.} \emph{et~al.}
\newblock \bibinfo{journal}{\bibinfo{title}{Diagnostic accuracy of digital
  screening mammography with and without computer-aided detection}}.
\newblock {\emph{\JournalTitle{JAMA internal medicine}}}
  \textbf{\bibinfo{volume}{175}}, \bibinfo{pages}{1828--1837}
  (\bibinfo{year}{2015}).

\bibitem{smith2005physician}
\bibinfo{author}{Smith-Bindman, R.} \emph{et~al.}
\newblock \bibinfo{journal}{\bibinfo{title}{Physician predictors of
  mammographic accuracy}}.
\newblock {\emph{\JournalTitle{Journal of the National Cancer Institute}}}
  \textbf{\bibinfo{volume}{97}}, \bibinfo{pages}{358--367}
  (\bibinfo{year}{2005}).

\bibitem{blanks1998comparison}
\bibinfo{author}{Blanks, R.}, \bibinfo{author}{Wallis, M.} \&
  \bibinfo{author}{Moss, S.}
\newblock \bibinfo{journal}{\bibinfo{title}{A comparison of cancer detection
  rates achieved by breast cancer screening programmes by number of readers,
  for one and two view mammography: results from the uk national health service
  breast screening programme}}.
\newblock {\emph{\JournalTitle{Journal of Medical screening}}}
  \textbf{\bibinfo{volume}{5}}, \bibinfo{pages}{195--201}
  (\bibinfo{year}{1998}).

\bibitem{karssemeijer2004effect}
\bibinfo{author}{Karssemeijer, N.}, \bibinfo{author}{Otten, J.~D.},
  \bibinfo{author}{Roelofs, A.~A.}, \bibinfo{author}{van Woudenberg, S.} \&
  \bibinfo{author}{Hendriks, J.~H.}
\newblock \bibinfo{title}{Effect of independent multiple reading of mammograms
  on detection performance}.
\newblock In \emph{\bibinfo{booktitle}{Medical Imaging 2004}},
  \bibinfo{pages}{82--89} (\bibinfo{organization}{International Society for
  Optics and Photonics}, \bibinfo{year}{2004}).

\bibitem{christoyianni2002computer}
\bibinfo{author}{Christoyianni, I.}, \bibinfo{author}{Koutras, A.},
  \bibinfo{author}{Dermatas, E.} \& \bibinfo{author}{Kokkinakis, G.}
\newblock \bibinfo{journal}{\bibinfo{title}{Computer aided diagnosis of breast
  cancer in digitized mammograms}}.
\newblock {\emph{\JournalTitle{Computerized medical imaging and graphics}}}
  \textbf{\bibinfo{volume}{26}}, \bibinfo{pages}{309--319}
  (\bibinfo{year}{2002}).

\bibitem{brem2003improvement}
\bibinfo{author}{Brem, R.~F.} \emph{et~al.}
\newblock \bibinfo{journal}{\bibinfo{title}{Improvement in sensitivity of
  screening mammography with computer-aided detection: a multiinstitutional
  trial}}.
\newblock {\emph{\JournalTitle{American Journal of Roentgenology}}}
  \textbf{\bibinfo{volume}{181}}, \bibinfo{pages}{687--693}
  (\bibinfo{year}{2003}).

\bibitem{ciatto2003comparison}
\bibinfo{author}{Ciatto, S.} \emph{et~al.}
\newblock \bibinfo{journal}{\bibinfo{title}{Comparison of standard reading and
  computer aided detection (cad) on a national proficiency test of screening
  mammography}}.
\newblock {\emph{\JournalTitle{European journal of radiology}}}
  \textbf{\bibinfo{volume}{45}}, \bibinfo{pages}{135--138}
  (\bibinfo{year}{2003}).

\bibitem{freer2001screening}
\bibinfo{author}{Freer, T.~W.} \& \bibinfo{author}{Ulissey, M.~J.}
\newblock \bibinfo{journal}{\bibinfo{title}{Screening mammography with
  computer-aided detection: prospective study of 12,860 patients in a community
  breast center}}.
\newblock {\emph{\JournalTitle{Radiology}}} \textbf{\bibinfo{volume}{220}},
  \bibinfo{pages}{781--786} (\bibinfo{year}{2001}).

\bibitem{morton2006screening}
\bibinfo{author}{Morton, M.~J.}, \bibinfo{author}{Whaley, D.~H.},
  \bibinfo{author}{Brandt, K.~R.} \& \bibinfo{author}{Amrami, K.~K.}
\newblock \bibinfo{journal}{\bibinfo{title}{Screening mammograms:
  interpretation with computer-aided detection—prospective evaluation}}.
\newblock {\emph{\JournalTitle{Radiology}}} \textbf{\bibinfo{volume}{239}},
  \bibinfo{pages}{375--383} (\bibinfo{year}{2006}).

\bibitem{warren2000potential}
\bibinfo{author}{Warren~Burhenne, L.~J.} \emph{et~al.}
\newblock \bibinfo{journal}{\bibinfo{title}{Potential contribution of
  computer-aided detection to the sensitivity of screening mammography 1}}.
\newblock {\emph{\JournalTitle{Radiology}}} \textbf{\bibinfo{volume}{215}},
  \bibinfo{pages}{554--562} (\bibinfo{year}{2000}).

\bibitem{gilbert2008single}
\bibinfo{author}{Gilbert, F.~J.} \emph{et~al.}
\newblock \bibinfo{journal}{\bibinfo{title}{Single reading with computer-aided
  detection for screening mammography}}.
\newblock {\emph{\JournalTitle{New England Journal of Medicine}}}
  \textbf{\bibinfo{volume}{359}}, \bibinfo{pages}{1675--1684}
  (\bibinfo{year}{2008}).

\bibitem{fenton2007influence}
\bibinfo{author}{Fenton, J.~J.} \emph{et~al.}
\newblock \bibinfo{journal}{\bibinfo{title}{Influence of computer-aided
  detection on performance of screening mammography}}.
\newblock {\emph{\JournalTitle{New England Journal of Medicine}}}
  \textbf{\bibinfo{volume}{356}}, \bibinfo{pages}{1399--1409}
  (\bibinfo{year}{2007}).

\bibitem{fenton2011effectiveness}
\bibinfo{author}{Fenton, J.~J.} \emph{et~al.}
\newblock \bibinfo{journal}{\bibinfo{title}{Effectiveness of computer-aided
  detection in community mammography practice}}.
\newblock {\emph{\JournalTitle{Journal of the National Cancer institute}}}
  \textbf{\bibinfo{volume}{103}}, \bibinfo{pages}{1152--1161}
  (\bibinfo{year}{2011}).

\bibitem{imagechecker10manual}
\bibinfo{author}{Hologic}.
\newblock \emph{\bibinfo{title}{Understanding ImageChecker® CAD 10.0 User
  Guide – MAN-03682 Rev 002}} (\bibinfo{year}{2017}).

\bibitem{hupse2009use}
\bibinfo{author}{Hupse, R.} \& \bibinfo{author}{Karssemeijer, N.}
\newblock \bibinfo{journal}{\bibinfo{title}{Use of normal tissue context in
  computer-aided detection of masses in mammograms}}.
\newblock {\emph{\JournalTitle{IEEE Transactions on Medical Imaging}}}
  \textbf{\bibinfo{volume}{28}}, \bibinfo{pages}{2033--2041}
  (\bibinfo{year}{2009}).

\bibitem{hupse2013standalone}
\bibinfo{author}{Hupse, R.} \emph{et~al.}
\newblock \bibinfo{journal}{\bibinfo{title}{Standalone computer-aided detection
  compared to radiologists’ performance for the detection of mammographic
  masses}}.
\newblock {\emph{\JournalTitle{European radiology}}}
  \textbf{\bibinfo{volume}{23}}, \bibinfo{pages}{93--100}
  (\bibinfo{year}{2013}).

\bibitem{kooi2017large}
\bibinfo{author}{Kooi, T.} \emph{et~al.}
\newblock \bibinfo{journal}{\bibinfo{title}{Large scale deep learning for
  computer aided detection of mammographic lesions}}.
\newblock {\emph{\JournalTitle{Medical image analysis}}}
  \textbf{\bibinfo{volume}{35}}, \bibinfo{pages}{303--312}
  (\bibinfo{year}{2017}).

\bibitem{krizhevsky2012imagenet}
\bibinfo{author}{Krizhevsky, A.}, \bibinfo{author}{Sutskever, I.} \&
  \bibinfo{author}{Hinton, G.~E.}
\newblock \bibinfo{title}{Imagenet classification with deep convolutional
  neural networks}.
\newblock In \emph{\bibinfo{booktitle}{Advances in neural information
  processing systems}}, \bibinfo{pages}{1097--1105} (\bibinfo{year}{2012}).

\bibitem{he2015delving}
\bibinfo{author}{He, K.}, \bibinfo{author}{Zhang, X.}, \bibinfo{author}{Ren,
  S.} \& \bibinfo{author}{Sun, J.}
\newblock \bibinfo{title}{Delving deep into rectifiers: Surpassing human-level
  performance on imagenet classification}.
\newblock In \emph{\bibinfo{booktitle}{Proceedings of the IEEE international
  conference on computer vision}}, \bibinfo{pages}{1026--1034}
  (\bibinfo{year}{2015}).

\bibitem{becker2017deep}
\bibinfo{author}{Becker, A.~S.} \emph{et~al.}
\newblock \bibinfo{journal}{\bibinfo{title}{Deep learning in mammography:
  Diagnostic accuracy of a multipurpose image analysis software in the
  detection of breast cancer.}}
\newblock {\emph{\JournalTitle{Investigative Radiology}}}
  (\bibinfo{year}{2017}).

\bibitem{dhungel2017fully}
\bibinfo{author}{Dhungel, N.}, \bibinfo{author}{Carneiro, G.} \&
  \bibinfo{author}{Bradley, A.~P.}
\newblock \bibinfo{title}{Fully automated classification of mammograms using
  deep residual neural networks}.
\newblock In \emph{\bibinfo{booktitle}{Biomedical Imaging (ISBI 2017), 2017
  IEEE 14th International Symposium on}}, \bibinfo{pages}{310--314}
  (\bibinfo{organization}{IEEE}, \bibinfo{year}{2017}).

\bibitem{lotter2017multi}
\bibinfo{author}{Lotter, W.}, \bibinfo{author}{Sorensen, G.} \&
  \bibinfo{author}{Cox, D.}
\newblock \bibinfo{title}{A multi-scale cnn and curriculum learning strategy
  for mammogram classification}.
\newblock In \emph{\bibinfo{booktitle}{Deep Learning in Medical Image Analysis
  and Multimodal Learning for Clinical Decision Support}},
  \bibinfo{pages}{169--177} (\bibinfo{publisher}{Springer},
  \bibinfo{year}{2017}).

\bibitem{dmchallenge}
\bibinfo{author}{DREAM}.
\newblock \bibinfo{title}{The digital mammography dream challenge}.
\newblock
  \bibinfo{howpublished}{\url{https://www.synapse.org/Digital\_Mammography\_DREAM\_challenge}}
  (\bibinfo{year}{2017}).

\bibitem{trister2017will}
\bibinfo{author}{Trister, A.~D.}, \bibinfo{author}{Buist, D.~S.} \&
  \bibinfo{author}{Lee, C.~I.}
\newblock \bibinfo{journal}{\bibinfo{title}{Will machine learning tip the
  balance in breast cancer screening?}}
\newblock {\emph{\JournalTitle{JAMA oncology}}}  (\bibinfo{year}{2017}).

\bibitem{heath2000digital}
\bibinfo{author}{Heath, M.}, \bibinfo{author}{Bowyer, K.},
  \bibinfo{author}{Kopans, D.}, \bibinfo{author}{Moore, R.} \&
  \bibinfo{author}{Kegelmeyer, W.~P.}
\newblock \bibinfo{title}{The digital database for screening mammography}.
\newblock In \emph{\bibinfo{booktitle}{Proceedings of the 5th international
  workshop on digital mammography}}, \bibinfo{pages}{212--218}
  (\bibinfo{organization}{Medical Physics Publishing}, \bibinfo{year}{2000}).

\bibitem{moreira2012inbreast}
\bibinfo{author}{Moreira, I.~C.} \emph{et~al.}
\newblock \bibinfo{journal}{\bibinfo{title}{Inbreast: toward a full-field
  digital mammographic database}}.
\newblock {\emph{\JournalTitle{Academic radiology}}}
  \textbf{\bibinfo{volume}{19}}, \bibinfo{pages}{236--248}
  (\bibinfo{year}{2012}).

\bibitem{ren2015faster}
\bibinfo{author}{Ren, S.}, \bibinfo{author}{He, K.}, \bibinfo{author}{Girshick,
  R.} \& \bibinfo{author}{Sun, J.}
\newblock \bibinfo{title}{Faster r-cnn: Towards real-time object detection with
  region proposal networks}.
\newblock In \emph{\bibinfo{booktitle}{Advances in neural information
  processing systems}}, \bibinfo{pages}{91--99} (\bibinfo{year}{2015}).

\bibitem{simonyan2014very}
\bibinfo{author}{Simonyan, K.} \& \bibinfo{author}{Zisserman, A.}
\newblock \bibinfo{journal}{\bibinfo{title}{Very deep convolutional networks
  for large-scale image recognition}}.
\newblock {\emph{\JournalTitle{arXiv preprint arXiv:1409.1556}}}
  (\bibinfo{year}{2014}).

\bibitem{jia2014caffe}
\bibinfo{author}{Jia, Y.} \emph{et~al.}
\newblock \bibinfo{journal}{\bibinfo{title}{Caffe: Convolutional architecture
  for fast feature embedding}}.
\newblock {\emph{\JournalTitle{arXiv preprint arXiv:1408.5093}}}
  (\bibinfo{year}{2014}).

\bibitem{bunch1977free}
\bibinfo{author}{Bunch, P.~C.}, \bibinfo{author}{Hamilton, J.~F.},
  \bibinfo{author}{Sanderson, G.~K.} \& \bibinfo{author}{Simmons, A.~H.}
\newblock \bibinfo{title}{A free response approach to the measurement and
  characterization of radiographic observer performance}.
\newblock In \emph{\bibinfo{booktitle}{Application of Optical Instrumentation
  in Medicine VI}}, \bibinfo{pages}{124--135}
  (\bibinfo{organization}{International Society for Optics and Photonics},
  \bibinfo{year}{1977}).

\bibitem{ellis2007evaluation}
\bibinfo{author}{Ellis, R.~L.}, \bibinfo{author}{Meade, A.~A.},
  \bibinfo{author}{Mathiason, M.~A.}, \bibinfo{author}{Willison, K.~M.} \&
  \bibinfo{author}{Logan-Young, W.}
\newblock \bibinfo{journal}{\bibinfo{title}{Evaluation of computer-aided
  detection systems in the detection of small invasive breast carcinoma}}.
\newblock {\emph{\JournalTitle{Radiology}}} \textbf{\bibinfo{volume}{245}},
  \bibinfo{pages}{88--94} (\bibinfo{year}{2007}).

\bibitem{sadaf2011performance}
\bibinfo{author}{Sadaf, A.}, \bibinfo{author}{Crystal, P.},
  \bibinfo{author}{Scaranelo, A.} \& \bibinfo{author}{Helbich, T.}
\newblock \bibinfo{journal}{\bibinfo{title}{Performance of computer-aided
  detection applied to full-field digital mammography in detection of breast
  cancers}}.
\newblock {\emph{\JournalTitle{European journal of radiology}}}
  \textbf{\bibinfo{volume}{77}}, \bibinfo{pages}{457--461}
  (\bibinfo{year}{2011}).

\end{thebibliography}

\section*{Author contributions statement}

D.R., I.C. and P.P. contributed to the conception and design of the study.
A.H. and Z.U. contributed to the acquisition, analysis and interpretation of data.
All authors reviewed the manuscript.

\section*{Additional information}
\subsection*{Competing financial interests}
The authors declare no competing financial interests.

\section*{Acknowledgements}

This work was supported by the Novo Nordisk Foundation Interdisciplinary Synergy Programme [Grant NNF15OC0016584]; and  National Research, Development and Innovation Fund of Hungary, [Project no. FIEK\_16-1-2016-0005].
The funding sources cover computational and publishing costs.
\end{document}